\documentclass[letterpaper]{article} 
\usepackage{aaai23}  
\usepackage{times}  
\usepackage{helvet}  
\usepackage{courier}  
\usepackage[hyphens]{url}  
\usepackage{graphicx} 
\usepackage{natbib}  
\usepackage{caption} 
\usepackage{tabularx}

\title{Computing Divergences between Discrete Decomposable Models}
\author{%
  Loong~Kuan~Lee,\textsuperscript{\rm 1}
  Nico~Piatkowski,\textsuperscript{\rm 2}
  François~Petitjean,\textsuperscript{\rm 1}
  and~Geoffrey~I.~Webb\textsuperscript{\rm 1}
}
\affiliations{
  \textsuperscript{\rm 1}
  Department of Data Science and AI, 
  Monash University, Melbourne, Australia\\
  \textsuperscript{\rm 2}Fraunhofer IAIS, 
  Schloss Birlinghoven, 53757 Sankt Augustin, Germany\\
  mail@lklee.dev
}

\usepackage{amssymb}
\usepackage{booktabs}
\usepackage[]{import}
\usepackage{adjustbox}
\usepackage{multicol}
\usepackage{multirow}
\usepackage[]{mathtools}
\usepackage{enumitem}
\usepackage{bm}
\usepackage{soul}

\usepackage{subfig}

\usepackage{color}

\usepackage{xcolor}
\newcommand\todo[1]{\colorbox{red}{(#1)}}

\usepackage{tikz}
\usetikzlibrary{positioning}
\usetikzlibrary{arrows.meta}
\usetikzlibrary{fit}
\usetikzlibrary{calc}

\usepackage{amsthm}
\newtheorem{theorem}{Theorem}
\newtheorem{definition}{Definition}
\newtheorem{remark}{Remark}

\let\svlim\lim\def\lim{\svlim\limits}

\let\iff\Leftrightarrow
\let\epsilon\varepsilon

\usepackage{stmaryrd} 

\newcommand{\vdom}[1]{\operatorname{Dom}(#1)}

\newcommand{\ab}{\alpha\beta}
\newcommand{\AB}{\text{AB}}
\newcommand{\jt}{\mathcal{T}}

\newcommand{\graph}{\mathcal{G}}
\newcommand{\graphH}{\mathcal{H}}
\newcommand{\graphP}{\mathcal{G}_\pr}
\newcommand{\graphQ}{\mathcal{G}_\qr}

\newcommand{\bigO}{\mathcal{O}}

\newcommand{\pnt}{\text{pa}}
\newcommand{\pa}{\text{pa}}

\newcommand{\tw}{\omega}
\newcommand{\SP}{\textit{SP}}

\newcommand{\clis}{\bm{\mathcal{C}}}
\newcommand{\seps}{\bm{\mathcal{S}}}
\newcommand{\model}{\pr_{\graph}}
\newcommand{\sep}{\mathcal{S}}
\newcommand{\cli}{\mathcal{C}}

\newcommand{\dr}{\mathbb{D}}
\newcommand{\pr}{\mathbb{P}}
\newcommand{\qr}{\mathbb{Q}}
\newcommand{\drjt}{\mathbb{D}^\jt}
\newcommand{\prjt}{\mathbb{P}^\jt}
\newcommand{\qrjt}{\mathbb{Q}^\jt}

\newcommand{\prejt}{\hat{\mathbb{P}}^\jt}
\newcommand{\qrejt}{\hat{\mathbb{Q}}^\jt}

\newcommand{\dom}{\mathcal{X}}

\newcommand{\func}{\mathcal{F}}
\newcommand{\xvar}{X}
\newcommand{\xvars}{\vec{X}}
\newcommand{\xval}{\vec{x}}
\newcommand{\modelP}{\pr_{\graphP}}
\newcommand{\modelQ}{\qr_{\graphQ}}
\let\vec\bm

\usepackage{glossaries}
\usepackage{glossaries-extra}

\glssetcategoryattribute{general}{nohyper}{true}
\glssetcategoryattribute{acronym}{nohyper}{true}

\setabbreviationstyle[acronym]{long-short}
\newglossaryentry{zg}{
  name={\texttt{zg}},
  description={method to estimate KL-divergence from \todo{zg}}
}
\newglossaryentry{hjw}{
  name={\texttt{hjw}},
  description={method to estimate KL-divergence from \todo{hjw}}
}
\newglossaryentry{bzlv}{
  name={\texttt{bzlv}},
  description={method to estimate KL-divergence from \todo{bzlv}}
}
\newglossaryentry{mcgo}{
  name={\texttt{mcgo}},
  description={method to compute KL-divergence from between \gls{bn} in~\cite{moral2021}}
}
\newglossaryentry{pgmpy}{
  name={\texttt{pgmpy}},
  description={Python library for graphical models~\cite{ankan2015}}
}
\newacronym{cpt}{CPT}{conditional probability table}
\newacronym{kl}{KL}{Kullback-Leibler}
\newacronym{bn}{BN}{Bayesian network}
\newacronym{mn}{MN}{Markov network}
\newacronym{dm}{DM}{decomposable model}
\newacronym{mle}{MLE}{maximum likelihood estimation}
\newacronym{jta}{JTA}{junction tree algorithm}
\newacronym{JTC}{JTComp}{Junction Tree Computation}
\newacronym{JFC}{JFComp}{Junction Forest Computation}
\newacronym{method}{DeMoDivEst}{decomposable model Divergence Estimator}
\newacronym{generator}{DeMoGen}{decomposable model Generator}

\nocopyright
\begin{document}

\maketitle

\begin{abstract}
  \sloppy
There are many applications that benefit from computing the exact divergence between 2 discrete probability measures, including machine learning. Unfortunately, in the absence of any assumptions on the structure or independencies within these distributions, computing the divergence between them is an intractable problem in high dimensions.
We show that we are able to compute a wide family of functionals and divergences, such as the alpha-beta divergence, between two decomposable models, i.e~chordal Markov networks, in time exponential to the treewidth of these models. The alpha-beta divergence is a family of divergences that include popular divergences such as the Kullback-Leibler divergence, the Hellinger distance, and the chi-squared divergence. Thus, we can accurately compute the exact values of any of this broad class of divergences to the extent to which we can accurately model the two distributions using decomposable models.


\end{abstract}

\section{Introduction}\label{sec:intro}
Computing the divergence, i.e.~the degree of ``difference'', between two joint probability distributions is a problem that has many applications in the field of Machine Learning. For instance, it can be used to estimate the divergence between the underlying distributions of two data samples.
This particular application is useful in the
study
of changing distributions, also known as concept drift {\cite{schlimmer1986,webb2018}, in the detection of anomalous regions in spatio-temporal data {\cite{barz2019,piatkowski2013}}, and in various tasks related to the retrieval, classification, and visualisation of time series data {\cite{chen2020}}.



  Although there has been much work in \emph{estimating} the divergence between 2 general \emph{high-dimensional} discrete distributions~\cite{bhattacharya2009,abdullah2016},
  they do not compute the exact divergence between these distributions as it is intractable to do so without any knowledge or assumptions made regarding the structure
  within these distributions.
Instead, approaches that do take advantage of some structural properties within the distributions for an efficient \emph{computation} of divergences have appeared in the literature before, e.g., for computing the \gls{kl} divergence between \glspl{bn}~\cite{moral2021}. It is also possible to tractably compute the \gls{kl} divergence between a general \gls{mn} and a \gls{mn} where inference tasks are tractable~\cite{koller2009}.


However, there are situations where one might want to compute divergences other than the \gls{kl} divergence \cite{Nowozin/etal/2016a}, in particular in the variational inference community where they have been employed to derive alternative evidence lower bounds \cite{Chen/etal/2018a,Li/Turner/2016a,Dieng/etal/2017a} or in the context of generative models \cite{Genevay/etal/2018a}.
Furthermore, in natural language processing, using the KL divergence is problematic in the presence of uneven word frequencies~\cite{Labeau/Cohen/2019a}.
Even for fundamental problems like model selection, we show that considering different types of divergences can be beneficial.



Motivated by these considerations, in this paper we show how to compute a wide family of divergences, the $\ab$-divergences, between two \glspl{dm}.
In the process of showing how the $\ab$-divergence can be computed between any two \glspl{dm}, we will reach a more general result. That is, we will show how one can compute, between two \glspl{dm}, the functional $\func$ defined in Definition~\ref{def:func}:
\begin{definition}\label{def:func}
  (Functional $\func$)
  \begin{align*}
    &\func(\pr,\qr; g, h, g^{*}, h^{*})\\
    &= \sum_{\xval \in \dom}
    \left[g\big[\pr\big](\xval)\right]
    \left[h\big[\qr\big](\xval)\right]
    L\Big(
    \left[g^{*}[\pr](\xval)\right]
    \left[h^{*}[\qr](\xval)\right]\Big)
    \end{align*}
    where, for any distribution $\pr$ defined by a \gls{dm} with graph structure $\graph$, $L$ is any function with the property $L\left(\prod_{r}r\right)=\sum_{r}L(r)$,
    and $f \in \{g, h, g^{*}, h^{*}\}$ are functionals with the property:
    \begin{gather}
      f\left[\prod_{\cli \in \clis(\graph)} \pr_{\cli} \right]
      (\xval_{\cli})
    =\prod_{\cli\in\clis(\graph)}f\big[\pr_{\cli}\big](\xval_{\cli})
    \label{eq:func-multi-prop}
  \end{gather}
\end{definition}

%
This result implies the possibility for the computation of divergences and functionals other than the $\ab$-divergence between two \glspl{dm}. In fact, we show that $\func$ can be computed by running the \gls{jta} over a specifically constructed chordal graph and set of initial factors.

%
Proofs for our contributed theoretical results are deferred to the technical appendix.


\section{Background and Notation}\label{sec:back}
Let us summarize the notation and background necessary for the subsequent development.




\subsection{Markov Networks (MNs)}\label{sec:back-mn}
\newcommand{\bX}{X}
\newcommand{\bt}{\boldsymbol{\theta}}
\newcommand{\mR}{\mathbb{R}}
\newcommand{\cD}{\mathcal{D}}

An undirected graph $\graph=(V,E)$ consists of $n=|V|$ vertices, connected via edges $(v,w)\in E$. For two graphs $\graph_1,\graph_2$, we write $V(\graph_1)$ and $V(\graph_2)$ to denote the vertices of $\graph_1$ and $\graph_2$, respectively and similar $E(\graph_1)$ and $E(\graph_2)$ for the edges. 
A clique $\cli$ is a fully-connected subset of vertices, i.e., $\forall v,w\in \cli : (v,w)\in E$. The set of all cliques of $\graph$ is denoted by $\clis(\graph)$. Here, any undirected graph represents the conditional independence structure of an undirected graphical model or \gls{mn} \cite{Wainwright/Jordan/2008a}.

To this end, we identify each vertex $v\in V$ with a random variable $X_v$ taking values in the state space $\dom_{v}=\vdom{X_v}$. The random vector $\xvars=(X_v : v\in V)$, with probability mass function (pmf) $\pr$, represents the random joint state of all vertices in some arbitrary but fixed order, taking values $\xval$ in the Cartesian product space $\dom=\vdom{\xvars}=\bigotimes_{v\in V} \dom_v$. If not stated otherwise, $\dom$ is a discrete set. Moreover, we allow to access these quantities for any proper subset of variables $S\subset V$, i.e., $\xvars_S = (X_v : v\in S)$, $\xval_S$, and $\dom_{S}=\bigotimes_{v\in S} \dom_v$, respectively. We write $\omega(\graph)$ to indicate the treewidth of $\graph$, i.e.~$\omega(\graph) = \max_{\cli \in \clis(\graph)} |\cli| - 1$.

According to the Hammersley-Clifford theorem \cite{Hammersley/Clifford/1971a}, the probability mass of $\xvars$ factorizes over positive functions $\psi_\cli : \dom \to \mR_+$, one for each maximal clique of the underlying graph,
\begin{equation}
\pr(\xvars=\xval) = \frac{1}{Z} \prod_{C\in\clis}\psi_\cli(\xval_\cli) \;,\label{eq:factorize}
\end{equation}
normalized via $Z=\sum_{\xval\in\dom}\prod_{\cli\in\clis}\psi_\cli(\xval_\cli)$. Due to positivity of $\psi_\cli$, it can be written as an exponential, i.e., $\psi_\cli(\xval_\cli)=\exp(\langle\bt_\cli,\phi_\cli(\xval_\cli)\rangle)$ with sufficient statistic $\phi_\cli : \dom_\cli\to\mR^{|\dom_\cli|}$. The overcomplete sufficient statistic of discrete data is a ``one-hot'' vector that selects a specific weight value, e.g., $\psi_\cli(\xval_\cli)=\exp(\bt_{\cli=\xval_\cli})$. The full joint can be written in the famous exponential family form $\pr(\xvars=\xval) = \exp(\langle\bt,\phi(\xval)\rangle - \log Z)$ with $\bt=(\bt_\cli : \cli\in\clis)$ and $\phi(\xval)=(\phi_\cli(\xval_\cli) : \cli\in\clis)$.

The parameters of exponential family members are estimated by
minimizing the negative average log-likelihood
$\ell(\bt;\cD)=-(1/|\cD|)\sum_{\xval\in\cD}\log\pr_{\bt}(\xval)$ for some
data set $\cD$ via first-order numeric optimization methods. $\cD$
contains samples from $\xvars$, and it can be shown that the estimated
probability mass converges to the data generating distribution as the
size of $\cD$ increases. However, computing $Z$ and hence performing
probabilistic inference is \#P-hard
\cite{Valiant/1979a,Bulatov/Grohe/2004a}. There are approximation techniques for inference with quality guarantees~\cite{piatkowski2018}, but for exact inference, the junction tree algorithm is needed. The junction tree
representation of an undirected model is a tree, in which each vertex
represents a maximal clique of a triangulation\footnote{A triangulation of a graph $\graph=(V,E)$ is another graph $\graph'=(V,E')$ with $E\subseteq E'$, such that $\graph'$ is a chordal graph.} of $\graph$
\cite[Sec.~2.5.2]{Wainwright/Jordan/2008a}. The cutset of each pair of
adjacent clique-vertices is called a separator.

Nevertheless, junction trees require the underlying graphical structure of the graphical model to be {\em decomposable}.

\subsection{Decomposable Models (DMs)}\label{sec:back-dm}
A {\em \gls{dm}}, $\model$, is a \gls{mn} where the underlying conditional independence structure, $\graph$, is a chordal graph
\footnote{A graph is cordial if every induced cycle has exactly 3 vertices.}.

\glspl{dm} can be translated directly into an equivalent junction tree representation by finding the {\em maximum spanning tree\/} of its {\em clique graph }. Each vertex of the clique graph is a maximal clique in the \gls{dm} and each edge is the separator between the vertex. The weight of each edge is then the number of variables in the corresponding separator.
The resulting junction tree, $\jt=(\clis, \seps)$, will have vertices that are the maximal cliques, $\clis$, and edges that are the minimal separators, $\seps$, of the \gls{dm}.

Beside allowing for fast inference, another benefit of a \gls{dm} is
that there is a closed form solution to the maximum likelihood parmeter estimation problem for the joint distribution over all the variables in the model \citep{haberman1977}.
Therefore, the joint distribution for the \gls{dm} $\model$ is:
%
\begin{align*}
 \model(\xval) &= \frac{\prod_{\cli \in \clis}\pr_{\cli}(\xval)}
 {\prod_{\sep \in \seps}\pr_{\sep}(\xval)}
\end{align*}
where $\pr_d(\cdot)$ represents the marginal probability over
$\dom_{d}$.

Alternatively, we can also represent the joint distribution of $\model$ as a product of \glspl{cpt} if we choose a maximal clique in $\clis$ to be the root node of $\model$'s junction tree $\jt$.
\begin{gather*}
  \model(\xval)
  = \prod_{\cli\in\clis}\pr_{\cli-\pa(\cli)|\pa(\cli)}
    (\xval_{\cli-\pa(\cli)} | \xval_{\pa(\cli)})
  =\prod_{\cli\in\clis} \pr^{\jt}_{\cli}(\xval)
\end{gather*}
where $\pr^{\jt}_{\cli}(\xval)=\pr_{\cli-\pa(\cli)|\pa(\cli)}(\xval_{\cli-\pa(\cli)} | \xval_{\pa(\cli)})$
and $\pa(\cli)$ is the parent clique of $\cli$ in the junction tree $\jt$. $\pa(\cli)=\emptyset$ when $\cli$ is the assigned root node of $\jt$.


\subsection{Junction Tree Algorithms (JTAs)}\label{sec:back-jta}
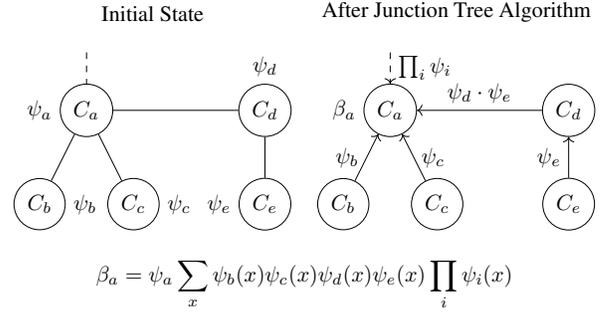
\begin{figure}
  \centering
  \begin{adjustbox}{width=0.45\textwidth}
    \begin{tikzpicture}[-,
  treenode/.style={circle, draw=black, fill=white, thin, minimum size=1mm},
  squarednode/.style={rectangle, draw=red!60, fill=red!5, very thick, minimum size=5mm},
  ]
  \node [treenode] (ca1) {$C_a$} {
    child { 
        node [treenode] (cb1) {$C_b$}
        edge from parent[draw] node [left] {}
    }
    child {
        node [treenode] (cc1) {$C_c$}
        edge from parent[draw] node [right] {}
    }
};
\node [treenode, right=20mm of ca1] (cd1) {$C_d$} {
    child { 
        node [treenode] (ce1) {$C_e$}
        edge from parent[draw] node [right] {}
    }
};
\node [above=5mm of ca1] (t1) {};
\draw [dashed] (t1) -- node [right]{} ++ (ca1);
\draw[draw] (ca1) -- (cd1);

  \node [left=0mm of ca1] (pa1) {$\psi_a$};
  \node [right=0mm of cb1] (pb1) {$\psi_b$};
  \node [right=0mm of cc1] (pc1) {$\psi_c$};
  \node [above=0mm of cd1] (pd1) {$\psi_d$};
  \node [left=0mm of ce1] (pe1){$\psi_e$};

  \node [
  fit=(pd1) (t1) (cd1) (cb1) (pc1)] (init) {};
  \node[above] at (init.north){Initial State};

  \node [right=1mm of ca1] (left) {};
  \node [right=16mm of ca1] (mid) {};

  \node [treenode, right=40mm of ca1] (ca2) {$C_a$} {
    child { 
        node [treenode] (cb2) {$C_b$}
        edge from parent[draw,<-] node [left] {$\psi_b$}
    }
    child {
        node [treenode] (cc2) {$C_c$}
        edge from parent[draw,<-] node [right] {$\psi_c$}
    }
};
\node [treenode, right=20mm of ca2] (cd2) {$C_d$} {
    child { 
        node [treenode] (ce2) {$C_e$}
        edge from parent[draw,<-] node [left] {$\psi_e$}
    }
};
\node [above=5mm of ca2] (t2) {};
\draw [dashed,->] (t2) -- node [right]{$\prod_i\psi_i$} ++ (ca2);
\draw[draw,<-] (ca2) -- node [above]{$\psi_d \cdot \psi_e$} ++ (cd2);

  \node [left=0mm of ca2] (pa2) {$\beta_a$};

  \node [
  fit=(t2) (cd2) (cc2) (cb2) (pa2) ] (fin) {};

  \node[above] at (fin.north)(fin2){After Junction Tree Algorithm};

  \node[below=2mm of {$(init.south) !.5! (fin.south)$}]{%
    $\begin{gathered}
      \beta_a=\psi_a\sum_x\psi_b(x)\psi_c(x)\psi_d(x)\psi_e(x)
      \prod_i\psi_i(x)
    \end{gathered}$};

\end{tikzpicture}

  \end{adjustbox}
  \caption{Illustration of the Junction Tree Algorithm.}\label{fig:jta}
\end{figure}

Recall the partition function of a general \gls{mn} $Z$:
$Z = \sum_{\vec{x}\in\dom} \prod_{\cli\in\clis}\psi_{\cli}(\vec{x}_{\cli})$. 
Evaluating the partition function of loopy models exactly does not necessarily require a naive summation over the state space $\dom$; there is another, more efficient, technique. Any loopy graph can be triangulated and converted into a chordal graph with a junction tree representation. \cite{Lauritzen/Spiegelhalter/1988a,Wainwright/Jordan/2008a}. Then, as illustrated in Figure~\ref{fig:jta}, the junction tree algorithm
goes a step further and computes the \emph{un-normalized} marginal ``probability'', $\beta$, for each maximal clique in the JT~\cite[Corollary~10.2]{koller2009}:\@
\begin{align*}
  \forall \cli \in \clis : \beta_{\cli}(\vec{x}_{\cli})
  = \sum_{\vec{x}\in\dom_{X - \cli}} \prod_{\cli\in\clis}
  \psi_{\cli}(\vec{x},\vec{x}_{\cli})
\end{align*}
As with belief propagation in ordinary trees, inference on the junction tree has a time complexity that is polynomial in the maximal state space size of its vertices. The maximal vertex state space size of a junction tree is, however, {exponential} in the treewidth of the triangulation of $\graph$ used. Hence, if the treewidth of the triangulation of a loopy model is small, exact inference via the junction tree algorithm is rather efficient. Choosing a triangulation that results in a minimal treewidth is an \textbf{NP}-hard problem, but a {valid} triangulation can be found with time and memory complexity linear in the number of vertices \cite{Dechter/2003a,Berry/etal/2004a,Heggernes/2006a}.
See Figure~\ref{fig:mrf} for an example of a non-minimal and minimal triangulation of a graph.

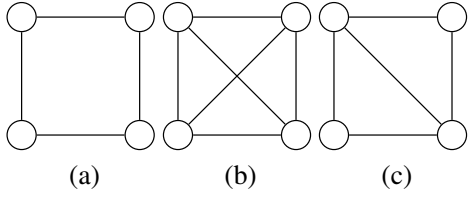
\begin{figure}
  \centering
  \begin{adjustbox}{width=0.35\textwidth}
    \subfloat[]{\begin{tikzpicture}[-,
  vert/.style={circle, draw=black, fill=white, thin, minimum size=1mm},
  squarednode/.style={rectangle, draw=red!60, fill=red!5, very thick, minimum size=5mm},
  ]
  \node [vert] (v1) {};
  \node [vert, right=10mm of v1] (v2) {};
  \node [vert, below=10mm of v1] (v3) {};
  \node [vert, right=10mm of v3] (v4) {};

  \draw[draw] (v1) edge (v2);
  \draw[draw] (v1) edge (v3);
  \draw[draw] (v4) edge (v2);
  \draw[draw] (v3) edge (v4);
\end{tikzpicture}

}
    \subfloat[]{\begin{tikzpicture}[-,
  vert/.style={circle, draw=black, fill=white, thin, minimum size=1mm},
  squarednode/.style={rectangle, draw=red!60, fill=red!5, very thick, minimum size=5mm},
  ]
  \node [vert] (v1) {};
  \node [vert, right=10mm of v1] (v2) {};
  \node [vert, below=10mm of v1] (v3) {};
  \node [vert, right=10mm of v3] (v4) {};

  \draw[draw] (v1) edge (v2);
  \draw[draw] (v1) edge (v3);
  \draw[draw] (v4) edge (v2);
  \draw[draw] (v3) edge (v4);
  \draw[draw] (v1) edge (v4);
  \draw[draw] (v3) edge (v2);
\end{tikzpicture}

}
    \subfloat[]{\begin{tikzpicture}[-,
  vert/.style={circle, draw=black, fill=white, thin, minimum size=1mm},
  squarednode/.style={rectangle, draw=red!60, fill=red!5, very thick, minimum size=5mm},
  ]
  \node [vert] (v1) {};
  \node [vert, right=10mm of v1] (v2) {};
  \node [vert, below=10mm of v1] (v3) {};
  \node [vert, right=10mm of v3] (v4) {};

  \draw[draw] (v1) edge (v2);
  \draw[draw] (v1) edge (v3);
  \draw[draw] (v4) edge (v2);
  \draw[draw] (v3) edge (v4);
  \draw[draw] (v1) edge (v4);
\end{tikzpicture}

}
  \end{adjustbox}
  \caption{(a) a non-chordal \gls{mn}, (b) a possible triangulation of this \gls{mn}, (c) a minimal triangulation of this \gls{mn}}\label{fig:mrf}
\end{figure}


\section{$\alpha\beta$-Divergence between \glspl{dm}}\label{sec:back-div}
A divergence is a measure of the ``difference'' between 2 probability distributions. More formally, a divergence is a function between 2 distributions as defined in Definition~\ref{def:divergence}.
\begin{definition}\label{def:divergence}
  (Divergence)
  Suppose $S$ is the set of probability distributions with the same support. A divergence, $D$, is the function
$D(\cdot\mid \mid\cdot) : S \times S \rightarrow \mathbb{R}$ such that $\forall \pr,\qr \in S : D(\pr\mid \mid\qr) \geq 0$ and $\pr = \qr \iff D(\pr\mid \mid\qr) = 0$~\footnote{Some authors also require that the quadratic part of the Taylor expansion of $D(p, p + dp)$ define a Riemannian metric on $S$~\cite{amari2016}. However, this requirement is not needed by the methods described in this paper and is therefore left out.}.
\end{definition}


Furthermore, there are also generalized divergences where common divergences, such as the \gls{kl} divergence, are special cases of the generalized divergence. Specifically, we will use the generalized divergence known as the $\alpha\beta$-divergence \cite{cichocki2011}.
\begin{definition}[$\alpha\beta$-divergence]\label{def:ab-divergence}
  The $\alpha\beta$-divergence, $D_{\AB}$, between 2 positive measures $\pr$ and $\qr$ is defined by the following, where $\alpha$ and $\beta$ are parameters:
  \begin{gather}
    D_{\AB}^{\alpha,\beta}(\pr,\qr)
    = \sum_{\xval\in\dom}d_{\AB}^{\alpha,\beta}\big(\pr(\xval),\qr(\xval)\big)
  \end{gather}
  where~\cite{cichocki2011}:
  \begin{align}
    &d_{\AB}^{(\alpha,\beta)}(\pr(\vec{x}), \qr(\vec{x}))\label{eq:AB-div}\\
    &=
      \begin{cases}
        -\frac{1}{\alpha\beta}\left(
          \pr(\vec{x})^{\alpha}\qr(\vec{x})^{\beta} -
          \frac{\alpha\pr(\vec{x})^{\alpha + \beta} }{\alpha + \beta} -
          \frac{\beta\qr(\vec{x})^{\alpha + \beta}}{\alpha + \beta} \right)
        &\\\hfill \text{for } \alpha,\beta,\alpha + \beta \neq 0\\
        \frac{1}{\alpha^{2}}\left(
          \pr(\vec{x})^{\alpha} \log \frac{\pr(\vec{x})^{\alpha}}{\qr(\vec{x})^{\alpha}} -
          \pr(\vec{x})^{\alpha} + \qr(\vec{x})^{\alpha}
        \right) &\\\hfill  \text{for } \alpha\neq 0,\beta=0\\
        \frac{1}{\alpha^{2}} \left(
          \log \frac{\qr(\vec{x})^{\alpha}}{\pr(\vec{x})^{\alpha}} +
          \left(\frac{\qr(\vec{x})^{\alpha}}{\pr(\vec{x})^{\alpha}}\right)^{-1} - 1
        \right) &\\\hfill  \text{for }\alpha=-\beta\neq 0\\
        \frac{1}{\beta^{2}}\left(
          \qr(\vec{x})^{\beta}\log\frac{\qr(\vec{x})^{\beta}}{\pr(\vec{x})^{\beta}} -
          \qr(\vec{x})^{\beta} + \pr(\vec{x})^{\beta}
        \right) &\\\hfill  \text{for } \alpha=0,\beta \neq 0\\
        \frac{1}{2}(\log \pr(\vec{x}) - \log \qr(\vec{x}))^{2}
        \hfill  \text{for } \alpha,\beta=0.
      \end{cases}\nonumber
    \end{align}
\end{definition}
%

%
The parameters $\alpha$ and $\beta$ in the $\alpha\beta$-divergence is used to express other commonly used divergences. Specifically, the $\alpha=1,\beta=0$ gives the \gls{kl} divergence, while the $\alpha=0.5,\beta=0.5$ gives the Bhattacharyya coefficient which immediately gives the Hellinger distance.


The expression of the $\alpha\beta$-divergence in Equation~\ref{eq:AB-div} can be expressed as a linear combination of 3 smaller functionals.
\begin{theorem}\label{thm:ab-div-f-expr}
  The 5 cases of the $\alpha\beta$-divergence in Equation~\ref{eq:AB-div} are linear combinations of the following 3 functionals:
  \begin{align*}
    f_{1}(\pr,\qr)
    &=\sum_{\xval\in\dom}\frac{1}{2}{\left(
      \log\pr(\xval) - \log\qr(\xval)
      \right)}^{2}\\
    f_{2}(\pr,\qr;a,b)
    &= \sum_{\xval\in\dom}\pr(\xval)^{a}\qr(\xval)^{b}\\
    f_{3}(\pr,\qr;a,b,c,d)
    &= \sum_{\xval\in\dom}\pr(\xval)^{a}\qr(\xval)^{b}
      \log(\pr(\xval)^{c}\qr(\xval)^{d})
  \end{align*}
\end{theorem}

%
Therefore, the ability to tractably compute these functionals between 2 \glspl{dm} will imply the ability to tractably compute the $\ab$-divergence between 2 \glspl{dm}.
Here we assume a complexity exponential to the treewidth of our \glspl{dm} is \emph{tractable}.
\begin{theorem}\label{thm:f1-timecmplx}
  The
  time complexity for computing the functional $f_{1}$ directly between 2 \glspl{dm} is $\mathcal{O}(n^{2} \tw 2^{\tw+1})$
  where $\tw(\graph)$ is the treewidth of chordal graph $\graph$, $\tw = \text{max}(\tw(\graphP), \tw(\graphQ))$, and $n$ is the number of variables.
\end{theorem}
Since computing $f_{1}$ directly is tractable, the focus of the rest of this paper will be to show how to compute functionals $f_{2}$ and $f_{3}$ between 2 \glspl{dm}.
In order to simplify further exposition, it will be ideal if functionals $f_{2}$ and $f_{3}$ can be expressed by a single, more general, functional.
%
\begin{theorem}\label{thm:express-f2}
  $f_{2}$ can be expressed by functional $\func$.
\end{theorem}
\begin{theorem}\label{thm:express-f3}
  $f_{3}$ can be expressed by functional $\func$.
\end{theorem}
Therefore, any method that can tractably compute $\func$, as defined in Definition~\ref{def:func}, between 2 \glspl{dm} can also tractably compute the $\ab$-divergence between these models.


With reasoning for the definition of $\func$ established, we can now substitute the maximum likelihood estimator of \glspl{dm} $\modelP$ and $\modelQ$ into functional $\func$. But before we start, first recall the notation established in Section~\ref{sec:back-dm}:
\begin{gather*}
  \pr(\xval)=\prod_{\cli\in\clis} \pr\left(\xval_{\cli - \pnt(\cli)} \mid \xval_{\pnt(\cli)}\right)=\prod_{\cli\in\clis}\prjt_{\cli}(\xval_{\cli}) \\
  \text{where, for } \cli \subset \xvar : \prjt_{\cli}\left(\xval_{\xvar}\right) = \prjt_{\cli}(\xval_{\cli})
\end{gather*}
and $\pnt(\cli)$ is the parent of the maximal clique $\cli$ in the junction tree of $\pr$'s and $\qr$'s respective chordal graph.
Then continuing with the substitution we get:
\begin{align}
  \func
  &({\pr},{\qr}; g, h, g^{*}, h^{*}
    )\label{eq:func-est-plug-in}\\
  =&\sum_{\xval\in\dom}
    \big(g\left[\pr\right](\xval)\big)
    \big(h\left[\qr\right](\xval)\big)
    L\left(
      \big(g^{*}\left[\pr\right](\xval)\big)
      \big(h^{*}\left[{\qr}\right](\xval)\big)
    \right)\nonumber\\
  =&\left[\sum_{\cli\in\clis_\pr}\sum_{\xval\in\dom}
      L\left(g^{*}\left[\prjt_{\cli}\right](\xval)\right)
      \big(g\left[{\pr}\right](\xval)\big)
      \big(h\left[{\qr}\right](\xval)\big)
    \right]+\nonumber\\
    &\left[\sum_{\cli\in\clis_\qr}\sum_{\xval\in\dom}
      L\left(h^{*}\left[\qrjt_{\cli}\right](\xval)\right)
      \big(g\left[{\pr}\right](\xval)\big)
      \big(h\left[{\qr}\right](\xval)\big)
    \right]\nonumber\\
    =&
    \sum_{\cli\in\clis(\graph_\pr)}\sum_{\xval_{\cli}\in\dom_{\cli}}
    L\left(g^{*}\left[\pr_{\cli}^{\jt}\right](\xval_{\cli})\right)
    \SP_{\cli}(\xval_{\cli})+\nonumber\\
    &\sum_{\cli\in\clis(\graph_\qr)}\sum_{\xval_{\cli}\in\dom_{\cli}}
    L\left(h^{*}\left[\qr_{\cli}^{\jt}\right](\xval_{\cli})\right)
    \SP_{\cli}(\xval_{\cli})\nonumber
\end{align}
where, for ease of notation:
\begin{gather}
\begin{aligned}
    &\SP_{\cli}(\xval_{\cli})\\
    &= \sum_{\xval \in \dom_{\xvar-\cli}}
    \Big(g\left[{\pr}\right](\xval_{\cli},\xval)\Big)
    \Big(h\left[{\qr}\right](\xval_{\cli},\xval)\Big)\\
    &=\sum_{\xval \in \dom_{\xvar-\cli}}\left[
      \prod_{\cli \in \clis_{\pr}} g\left[\pr_{\cli}^{\jt}\right](\xval)
    \right]
    \left[
      \prod_{\cli \in \clis_{\qr}} h\left[\qr_{\cli}^{\jt}\right](\xval)
    \right]
\end{aligned}\label{eq:sp}
\end{gather}
which represents the marginalisation of all the variables that are not in the clique $\cli$ over the all the non-log factors produced by $\func$.
The equality in Equation~\ref{eq:func-est-plug-in}
holds mainly due to the associativity of summations.
%
\begin{remark}\label{prop:naive-complex}
  The lower bound complexity of directly computing
  Equation~\ref{eq:func-est-plug-in} is $\Omega(2^n)$
  where $n$ is the number of variables.
  Therefore directly computing the functional $\func$ between 2 \glspl{dm} is intractable.
\end{remark}
%
%
Consequently, in order to compute $\func(\pr, \qr)$, and therefore $D_{\AB}(\pr, \qr)$, while avoiding complexity exponential to $n$, we require a more sophisticated method for its computation.
%

\section{Computing Functional $\func$ between \glspl{dm}}\label{sec:compute}
In order to tractably compute $\func$ between \glspl{dm} $\modelP$ and $\modelQ$, and therefore the $\ab$-divergence, we first require knowledge of a {\em computation graph\/} between \glspl{dm} $\modelP$ and $\modelQ$.

\begin{definition}[strictly larger, clique mapping $\alpha$]
  \label{def:strictly-larger}\label{def:cli-map}
  A chordal graph $\graphH$ is \textit{strictly larger} than chordal
  graphs $\graph_{\pr}$ and $\graph_{\qr}$ if all the maximal cliques
  in both chordal graphs is either a subset or equal to a maximal
  clique in $\graphH$.
  In other words, $\graphH$ is \textit{strictly larger} than
  $\graph_{\pr}$ and $\graph_{\qr}$ if and only if there exists a
  mapping $\alpha$ such that:
  \begin{align*}
    &\alpha : \clis(\graph_\pr, \graph_\qr) \rightarrow \clis(\graphH)\\
    &\text{s.t.} \quad \forall \cli \in \clis(\graph_\pr, \graph_\qr) : \cli \subseteq \alpha(\cli)
  \end{align*}
  where
  $\clis(\graph_\pr, \graph_\qr) = \clis(\graph_\pr)\cup\clis(\graph_\qr)$
  is the set of maximal cliques in chordal graphs $\graph_\pr$ and
  $\graph_\qr$.
\end{definition}
\begin{definition}[computation graph]\label{def:comp-graph}
  If a chordal graph, $\graphH$, is \textit{strictly larger} than
  chordal graphs $\graph_{\pr}$ and $\graph_{\qr}$, then $\graphH$ is
  a \textit{computation graph} of \glspl{dm}
  $\modelP$ and
  $\modelQ$.
\end{definition}
We can obtain the computation graph $\graphH$ by first taking the graph union of $\graphP$ and $\graphQ$, and then triangulating $\graphP \cup \graphQ$.
%
%

For the rest of this section, we will provide details on how our method, \gls{JFC}, uses the \textit{junction tree algorithm} to compute the functional $\func$
between 2 \glspl{dm}.
We will first describe how \gls{JFC} works on a connected computation graph $\graphH$ before generalising this to cases when $\graphH$ is a disconnected graph.
%

\begin{figure}
  \centering
  \begin{adjustbox}{width=0.47\textwidth}
    \begin{tikzpicture}[-,
  treenode/.style={circle, draw=black, fill=white, thin, minimum size=1mm},
  squarednode/.style={rectangle, draw=red!60, fill=red!5, very thick, minimum size=5mm},
  ]
  \node [treenode] (ca1) {$C_a$} {
    child { 
        node [treenode] (cb1) {$C_b$}
        edge from parent[draw] node [left] {}
    }
    child {
        node [treenode] (cc1) {$C_c$}
        edge from parent[draw] node [right] {}
    }
};
\node [treenode, right=20mm of ca1] (cd1) {$C_d$} {
    child { 
        node [treenode] (ce1) {$C_e$}
        edge from parent[draw] node [right] {}
    }
};
\node [above=5mm of ca1] (t1) {};
\draw [dashed] (t1) -- node [right]{} ++ (ca1);
\draw[draw] (ca1) -- (cd1);

  \node [left=0mm of ca1] (pa1) {$\psi_a$};
  \node [right=0mm of cb1] (pb1) {$\psi_b$};
  \node [right=0mm of cc1] (pc1) {$\psi_c$};
  \node [above=0mm of cd1] (pd1) {$\psi_d$};
  \node [left=0mm of ce1] (pe1){$\psi_e$};

  \node [
  fit=(pd1) (t1) (cd1) (cb1) (pc1)] (init) {};
  \node[above] at (init.north){Initial State};

  \node [right=1mm of ca1] (left) {};
  \node [right=16mm of ca1] (mid) {};

  \node [treenode, right=40mm of ca1] (ca2) {$C_a$} {
    child { 
        node [treenode] (cb2) {$C_b$}
        edge from parent[draw,<-] node [left] {$\psi_b$}
    }
    child {
        node [treenode] (cc2) {$C_c$}
        edge from parent[draw,<-] node [right] {$\psi_c$}
    }
};
\node [treenode, right=20mm of ca2] (cd2) {$C_d$} {
    child { 
        node [treenode] (ce2) {$C_e$}
        edge from parent[draw,<-] node [left] {$\psi_e$}
    }
};
\node [above=5mm of ca2] (t2) {};
\draw [dashed,->] (t2) -- node [right]{$\prod_i\psi_i$} ++ (ca2);
\draw[draw,<-] (ca2) -- node [above]{$\psi_d \cdot \psi_e$} ++ (cd2);

  \node [left=0mm of ca2] (pa2) {$\beta_a$};

  \node [
  fit=(t2) (cd2) (cc2) (cb2) (pa2) ] (fin) {};

  \node[above] at (fin.north)(fin2){After Junction Tree Algorithm};

  \node[below=2mm of {$(init.south) !.5! (fin.south)$}]{%
    $\begin{gathered}
      \psi_i=\prod_{C_\mathbb{P}\in{A}_i(\bm{\mathcal{C}}(\mathcal{G}_\mathbb{P}))}
      g\left({\mathbb{P}}^{\mathcal{ T }}_{C_\mathbb{P}}\right)
      \prod_{C_\mathbb{Q}\in{A}_i(\bm{\mathcal{C}}(\mathcal{G}_\mathbb{Q}))}
      h\left({\mathbb{Q}}^{\mathcal{ T }}_{C_\mathbb{Q}}\right)\\
      \text{where}: A_{i}(\clis) = \{\cli : \cli \in \clis \wedge \alpha(\cli)==\cli_{i}\}\\
      \beta_a=\psi_a\sum_{\xval\in\dom}\psi_b(\xval)\psi_c(\xval)\psi_d(\xval)\psi_e(\xval)
      \prod_i\psi_i(\xval)
    \end{gathered}$};

\end{tikzpicture}

  \end{adjustbox}
  \caption{Junction Tree Algorithm to compute the functional $\func$
    between 2 \glspl{dm} $\modelP$ and $\modelQ$ using
    computation graph $\graphH$, assuming $\graphH$ is a connected graph.
  }\label{fig:jtc}
\end{figure}


\subsection{\acrfull{JTC}}\label{sec:comp-jtc}
Recall we want to compute $\func(\pr,\qr)$
as expressed in Equation~\ref{eq:func-est-plug-in}.
%
%
Observe that the 2 nested sums in $\func(\pr,\qr)$
has innermost sums over $\dom_{\cli}$ with
similar forms but over different sets of maximal cliques,
$\clis(\graph_{\pr})$ and $\clis(\graph_{\qr})$ respectively.
Therefore, we want to re-express
$\func(\pr,\qr)$ such that the two sums over $\dom_{\cli}$
are over the same set of maximal cliques, $\clis(\graphH)$.
%
\begin{theorem}\label{thm:sp-eq}
  Assume we have 2 \glspl{dm} $\modelP$ and $\modelQ$ and a
  computation graph $\graphH$ for both models.
  By Definition~\ref{def:cli-map}, we also have a mapping $\alpha$ from
  maximal cliques in $\modelP$ and $\modelQ$ to maximal cliques in
  $\graphH$.
  Then the following equivalences holds for $\dr\in\{\pr,\qr\}$:
  \begin{align}
      &\sum_{\cli\in\clis(\graph_\dr)}\sum_{\xval_{\cli}\in\dom_{\cli}}
        L\left(g^{*}\left[\drjt_{\cli}\right](\xval_{\cli})\right)
        \SP_{\cli}(\xval_{\cli})\label{eq:thm-sp-eq-d}\\
      &=\sum_{\cli\in\clis(\graph_\dr)}
      \sum_{\substack{\xval_{\alpha(\cli)}\in\\\dom_{\alpha(\cli)}}}
      L\left(g^{*}\left[\drjt_{\cli}\right](\xval_{\alpha(\cli)})\right)
      \SP_{\alpha(\cli)}(\xval_{\alpha(\cli)})\nonumber
  \end{align}
\end{theorem}
%
%
With that, we will now show how the computation of $\SP_{\alpha(\cli)}, \forall \cli\in\clis(\graphH)$ is equivalent to the running the \emph{junction tree algorithm} over the junction tree of $\graphH$ with a set of specific initial factors.
%
Figure~\ref{fig:jtc} shows an illustration of this procedure that we will now describe.
%
\begin{theorem}\label{thm:sp-compute}
  Given
  2 \glspl{dm}, $\modelP$ and
  $\modelQ$, and a
  computation graph of both models,
  $\graphH$.
  Let $\Psi$ be a set of factors defined as follows:
  \begin{align*}
    \Psi
    := &\left\{g\circ\prjt_{\cli_\pr} :
         \cli_\pr \in \clis(\graph_{\pr}\right)\} \bigcup
      \left\{h\circ \qrjt_{\cli_\qr} :
      \cli_\qr \in \clis(\graph_{\qr})\right\}
  \end{align*}
  After running the junction tree algorithm over the junction tree of
  $\graphH$ with factors $\Psi$, we will get the following
  beliefs over each maximal clique in $\graphH$:
  \begin{align*}
    \forall \cli \in \clis(\graphH) :
    \beta_{\cli}(\xval_{\cli})
    &= \SP_{\cli}(\xval_{\cli})
  \end{align*}
\end{theorem}
Therefore, using the junction tree algorithm,
we obtain beliefs over each maximal
clique in $\graphH$ that we can use to substitute for
$\SP$ in Equation~\ref{eq:func-est-plug-in}.
However, this assumes that $\graphH$ is a connected graph, which might not always be the case.

\subsection{\acrfull{JFC}}\label{sec:comp-jfc}
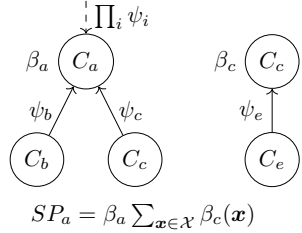
\begin{figure}
  \centering
  \begin{adjustbox}{width=0.47\textwidth}
    \begin{tikzpicture}[-,
treenode/.style={circle, draw=black, fill=white, thin, minimum size=1mm},
squarednode/.style={rectangle, draw=red!60, fill=red!5, very thick, minimum size=5mm},
]
\node [treenode] (ca1) {$C_a$} {
    child { 
        node [treenode] (cb1) {$C_b$}
        edge from parent[draw,<-] node [left] {$\psi_b$}
    }
    child {
        node [treenode] (cc1) {$C_c$}
        edge from parent[draw,<-] node [right] {$\psi_c$}
    }
};
\node [treenode, right=20mm of ca1] (cd1) {$C_d$} {
    child { 
        node [treenode] (ce1) {$C_e$}
        edge from parent[draw,<-] node [left] {$\psi_e$}
    }
};
\node [above=5mm of ca1] (t1) {};
\draw [dashed,->] (t1) -- node [right]{$\prod_i\psi_i$} ++ (ca1);
\draw[draw,<-] (ca1) -- node [above]{$\psi_d \cdot \psi_e$} ++ (cd1);

\node [left=0mm of ca1] (ba1) {$\beta_a$};

\node[fit=(ba1) (ce1) (ca1) (cb1) (t1)] (jt) {};

\node[below] at (jt.south) (eq1) {%
$\begin{aligned}
\text{SP}_{a} = \beta_{a}
\end{aligned}$};
\node[
fit=(ba1) (ce1) (ca1) (cb1) (t1)] (all) {};
\node[above right] at (all.north west){Junction Tree Computation};

\node [treenode, right=40mm of ca1] (ca2) {$C_a$} {
    child { 
        node [treenode] (cb2) {$C_b$}
        edge from parent[draw,<-] node [left] {$\psi_b$}
    }
    child {
        node [treenode] (cc2) {$C_c$}
        edge from parent[draw,<-] node [right] {$\psi_c$}
    }
};
\node [treenode, right=20mm of ca2] (cc2) {$C_c$} {
    child { 
        node [treenode] (ce2) {$C_e$}
        edge from parent[draw,<-] node [left] {$\psi_e$}
    }
};
\node [above=5mm of ca2] (t2) {};
\draw [dashed,->] (t2) -- node [right]{$\prod_i\psi_i$} ++ (ca2);

\node [left=0mm of ca2] (ba2) {$\beta_a$};
\node [left=0mm of cc2] (bc2) {$\beta_c$};

\node[fit=(bc2) (ce2) (ca2) (cb2) (t2)] (jt) {};
\node[above right] at (jt.north west){Junction Forest Computation};

\node[below] at (jt.south) (eq2) {%
$\begin{aligned}
&SP_a = \beta_a \textstyle\sum_{\xval\in\dom} \beta_c(\xval)
\end{aligned}$};
\node[
fit=(bc2) (ce2) (ca2) (cb2) (t2) ] (all2) {};

\end{tikzpicture}
  \end{adjustbox}
  \caption{Differences in getting the clique beliefs over each maximal
  clique in the computation graph $\graphH$ between a connected
  and a disconnected computation graph}\label{fig:jfc}
\end{figure}
We now show how \gls{JFC} can be extended to handle cases where
$\graphH$ is disconnected.

When the computation graph $\graphH$ is disconnected, $\graphH$ can be
represented as a list of chordal graphs, $\graphH=\{\graphH_{i}\}$.
Therefore, we also have a list of clique trees for each chordal
graph in $\graphH$, $\jt=\{\jt_{i}\}$, as well.

Since the
$\graphH$ is still strictly
larger than the chordal graph structure of $\modelP$ and $\modelQ$,
by Definition~\ref{def:cli-map}, there is still a mapping $\alpha$ from
$\clis(\graphP)\cup\clis(\graphQ)$ to $\clis(\graphH)$.
However, since chordal graph $\graphH$ is now comprised of multiple
chordal graphs, and therefore clique trees, we are unable to apply
Theorem~\ref{thm:sp-compute} directly to compute $\SP$ from
Equation~\ref{eq:sp}.
The reason for this is because there is no single clique tree to run
the junction tree algorithm on, therefore factors from different
clique trees are unable to propagate to each other.

Instead, we show in Theorem~\ref{thm:jfc-compute}, that having a
disconnected computation graph $\graphH$, and therefore a set of
clique trees, $\jt$, which are disconnected from each other,
essentially breaks up $\SP$ into smaller sub-problems over each
clique tree in $\jt$.
The results of these sub-problems can then be combined via multiplication to compute $\SP$.
An illustration of the result from Theorem~\ref{thm:jfc-compute} and
its difference in computing $\SP$ on a connected $\graphH$ can be found in Figure~\ref{fig:jfc}.

\begin{definition}[$\tau$, clique to clique tree mapping]\label{def:tree-map}

  Let $\tau$ be a mapping from the maximal cliques of $\modelP$ and
  $\modelQ$ to a clique tree in $\jt$ that contains the maximal
  clique given by the clique mapping, $\alpha$, from
  Definition~\ref{def:cli-map}:
  \begin{align*}
    &\tau : \clis(\graph_{\pr}) \cup \clis(\graph_{\qr}) \rightarrow \jt\\
    \text{s.t.} \quad
    &\forall \cli \in \clis(\graph_{\pr}) \cup \clis(\graph_{\qr}) :
      \alpha(\cli) \in \clis(\tau(\cli))
  \end{align*}
\end{definition}
\begin{theorem}\label{thm:jfc-compute}
  If the computation graph $\graphH$ for \glspl{dm} $\modelP$
  and $\modelQ$ is disconnected, $SP_{\alpha(\cli)}$ in
  Equation~\ref{eq:thm-sp-eq-d} can be
  re-expressed as follows:
  \begin{align*}
    &SP_{\alpha(\cli)}(\xval_{\alpha(\cli)})\\
    &=\beta_{\alpha(\cli)}(\xval_{\alpha(\cli)})
      \prod_{\jt_{i} \in \jt - \tau(\cli)} \sum_{\xval\in \dom_{\cli(\jt_{i})}}
      \beta_{\cli(\jt_{i})}(\xval_{\alpha(\cli)},\xval)\\
    &=\beta_{\alpha(\cli)}(\xval_{\alpha(\cli)}) R_{\tau(\cli)}
  \end{align*}
  where $\cli(\jt_{i})$ represents any clique in the set of
  maximal cliques in clique tree $\jt_{i}$ and $R_{\tau(\cli)}\in\mathbb{R}$.
\end{theorem}
%
%
Then we can obtain the required beliefs by running the junction tree algorithm for each junction tree in $\jt$ separately.
Therefore, even if the computation graph $\graphH$ is disconnected, we can compute $\SP$ and thus $\func(\pr,\qr)$.
Substituting these beliefs back into Equation~\ref{eq:func-est-plug-in} we get:
\begin{align}
  &\func(
  {\pr},{\qr})\label{eq:func-plug-in-final}\\
  &=
     \sum_{\cli\in\clis(\graph_\pr)}
    R_{\tau(\cli)}
     \sum_{\substack{\xval_{\alpha(\cli)}\in\\\dom_{\alpha(\cli)}}}
     L\left(g^{*}\left[\prjt_{\cli}\right](\xval_{\alpha(\cli)})\right)
  \beta_{\alpha(\cli)}(\xval_{\alpha(\cli)})\nonumber\\
  &+
    \sum_{C\in\clis(\graph_\qr)}
    R_{\tau(\cli)}
    \sum_{\substack{\xval_{\alpha(\cli)}\in\\\dom_{\alpha(\cli)}}}
     L\left(h^{*}\left[\qrjt_{\cli}\right](\xval_{\alpha(\cli)})\right)
  \beta_{\alpha(\cli)}(\xval_{\alpha(\cli)})\nonumber
\end{align}
Section~\ref{sec:comp-cmplx} will show that the complexity of computing the expression in Equation~\ref{eq:func-plug-in-final}, and in general, that the complexity of \gls{JFC} is more efficient than computing $\func$ directly.
%


\section{Computational Complexity}\label{sec:comp-cmplx}
We can determine the computational complexity of computing the $\ab$-divergence between 2 \glspl{dm} by first checking what the given values for $\alpha$ and $\beta$ are. This step takes $\mathcal{O}(1)$ time.
When $\alpha,\beta=0$, from Theorem~\ref{thm:f1-timecmplx} we know that the complexity of computing $D_{AB}^{0,0}(\pr\mid \mid \qr)$ is:
\begin{gather*}
  D_{AB}^{0,0}(\pr,\qr) \in \mathcal{O}(n^{2} \tw 2^{\tw+1})
\end{gather*}
where $\tw(\graph)$ is the treewidth of chordal graph $\graph$ and $\tw = \text{max}(\tw(\graphP), \tw(\graphQ))$.

When $\alpha$ and $\beta$ takes values other than $0$, we require the use of \gls{JFC} to compute parameterisations of $\func$ between $\modelP$ and $\modelQ$. In general, the $\ab$-divergence is a linear combination of different parameterisations of $\func$. Therefore, the complexity of computing the $\ab$-divergence is equivalent to computing $\func$ in big-O notation. As such, for the remainder of this section, we will discuss the overall complexity of \gls{JFC} for computing $\func$ between 2 \glspl{dm}.

The first step of \gls{JFC} involves assigning factors constructed from the \glspl{cpt} over
$\clis(\graphP)$ and $\clis(\graphQ)$ to $\clis(\graphH)$
Therefore, for each factor $\psi$, and therefore for each
$\cli \in \clis(\graphP) \cup \clis(\graphQ)$,
we need to search through $\clis(\graphH)$
to find a suitable clique to assign $\psi$ to. This results in the complexity:
\begin{equation*}
  \mathcal{O}(|\clis(\graphP)|\cdot |\clis(\graphH)|) + \mathcal{O}(|\clis(\graphP)|\cdot |\clis(\graphH)|)
  \in \mathcal{O}(n^{2})
\end{equation*}
since the number of maximal cliques in any chordal graph is bounded by the number of vertices in the graph.

Once all $\psi$s have been assigned to a maximal clique in $\graphH$, we then run the junction tree algorithm to calibrate the clique tree(s) of $\graphH$ with these factors. The complexity of this is:
\begin{equation*}
  \mathcal{O}(|\clis(\graphH)| \cdot 2^{\tw(\graphH)+1})
  \in
  \mathcal{O}(n \cdot 2^{\tw(\graphH)+1})
\end{equation*}
Once the clique tree/forest is calibrated and we know $\beta_{\alpha(\cli)}$ for all $\cli \in \clis(\graphP) \cup \clis(\graphQ)$, we can then compute Equation~\ref{eq:func-plug-in-final}:
\begin{align*}
  &\func(
  {\pr},{\qr})\nonumber\\
  &=
     \sum_{\cli\in\clis(\graph_\pr)}
    R_{\tau(\cli)}
     \sum_{\substack{\xval_{\alpha(\cli)}\in\\\dom_{\alpha(\cli)}}}
     L\left(g^{*}\left[\prjt_{\cli}\right](\xval_{\alpha(\cli)})\right)
  \beta_{\alpha(\cli)}(\xval_{\alpha(\cli)})\\
  &+
    \sum_{C\in\clis(\graph_\qr)}
    R_{\tau(\cli)}
    \sum_{\substack{\xval_{\alpha(\cli)}\in\\\dom_{\alpha(\cli)}}}
     L\left(h^{*}\left[\qrjt_{\cli}\right](\xval_{\alpha(\cli)})\right)
  \beta_{\alpha(\cli)}(\xval_{\alpha(\cli)})\\
  &\in \mathcal{O}(\clis(\graphP) \cdot 2^{\tw(\graphH)+1}) +
    \mathcal{O}(\clis(\graphQ) \cdot 2^{\tw(\graphH)+1})\\
  &\in\mathcal{O}(n \cdot 2^{\tw(\graphH)+1})
\end{align*}
Adding up the computational complexity of each step in \gls{JFC} results in the final complexity of computing the functional $\func$ between 2 \glspl{dm} $\modelP$ and $\modelQ$:
\begin{align*}
  &\mathcal{O}(n^{2}) +
  \mathcal{O}(n\cdot 2^{\tw(\graphH)+1}) +
  \mathcal{O}(n\cdot 2^{\tw(\graphH)+1})\\
  &\in\mathcal{O}(n\cdot 2^{\tw(\graphH)+1})
\end{align*}
which is more efficient than $\Omega(2^{n})$, the complexity of computing $\func$ directly.

Therefore, the computational complexity of computing the $\ab$-divergence between $\modelP$ and $\modelQ$ is:
\begin{align*}
  D_{AB}^{(\alpha,\beta)}(\pr\mid \mid \qr)
  \in
  \begin{cases}
    \mathcal{O}(n^{2} \cdot \tw 2^{\tw+1}) & \alpha,\beta=0\\
    \mathcal{O}(n\cdot 2^{\tw(\graphH)+1}) & \text{otherwise}
  \end{cases}
\end{align*}


\section{Runtime Comparison with \gls{mcgo}}\label{sec:runtime}
\begin{table}
  \centering
  \begin{tabularx}{\columnwidth}{lcccc}
    \toprule
    Network & \multicolumn{2}{c}{\gls{mcgo} (secs)} & \multicolumn{2}{c}{\gls{JFC} (secs)}\\
    & mean & sd & mean & sd \\
    \midrule
    cancer&\textbf{0.0117}&0.0026&0.0132&0.0033\\
    earthquake&0.0104&0.0025&\textbf{0.0075}&0.0009\\
    survey&0.0140&0.0032&\textbf{0.0081}&0.0002\\
    asia&0.0163&0.0001&\textbf{0.0137}&0.0007\\
    sachs&0.0464&0.0106&\textbf{0.0151}&0.0001\\
    child&0.0778&0.0101&\textbf{0.0402}&0.0013\\
    insurance&0.3838&0.0051&\textbf{0.1590}&0.0029\\
    water&\textbf{6.9326}&0.0329&7.6454&0.0637\\
    mildew&\textbf{19.326}&0.1318&19.459&0.0852\\
    alarm&0.3177&0.0099&\textbf{0.0875}&0.0018\\
    hailfinder&0.8543&0.0243&\textbf{0.1672}&0.0052\\
    hepar2&1.3058&0.0307&\textbf{0.2403}&0.0140\\
    win95pts&1.0256&0.0289&\textbf{0.3538}&0.0049\\
    \bottomrule
  \end{tabularx}
  \caption{Mean runtimes in seconds and their standard deviation for \gls{mcgo} and \gls{JFC} on computing the KL divergence between 2 \gls{bn}. The lower the better. Fastest times are bold.}\label{tbl:runtime}
\end{table}
Recall that a method already exists for computing the \gls{kl} divergence between 2 \glspl{bn}~\cite{moral2021} which we will refer to as \gls{mcgo}. Also note that it is possible to take a distribution represented by a \gls{bn} and, in exchange for some loss in independence information, represent it using a \gls{dm} instead~\cite[p.p.~134]{koller2009}. Therefore,
one might ask,
how does the practical runtime of \gls{JFC} compare to \gls{mcgo} 
when
computing the \gls{kl} divergence between 2 \glspl{bn}.

To answer this question, we will replicate the experiment used by~\citeauthor{moral2021}.
They chose a set of \glspl{bn} from the \textit{bnlearn}~\cite{scutari2010} repository (\url{https://www.bnlearn.com/bnrepository/}) to sample from and estimated a second \gls{bn} from these samples.
The authors have provided these \emph{estimated} \glspl{bn} for each of the \gls{bn} from \textit{bnlearn} used in their experiments: \url{https://github.com/mgomez-olmedo/KL-pgmpy}. Therefore, we will use this set of \glspl{bn} from their repository in our own experiments.

Now that we have
multiple pairs of \glspl{bn},
one original and one estimated from samples, we then compute the \gls{kl} divergence between each \gls{bn} pair using both \gls{mcgo} and \gls{JFC}.
We repeat this $10$ times
in order to get an estimate of both methods' runtime in seconds. We also do not factor in the conversion of these \glspl{bn} into \glspl{dm} in the final runtime.


We run the experiments
on an Intel NUC-10i7FNH with 64GB of RAM. The implementation of both methods are in Python and use the \gls{pgmpy} library~\cite{ankan2015}. The repository for the implementation for \gls{JFC} is \url{https://gitlab.com/lklee/comp-div-dm}. The results on the ``barley'' network for \gls{mcgo} is missing due to a lack of available memory on the system.

From the results in Table~\ref{tbl:runtime}, we can observe that despite \gls{mcgo} containing numerous computation optimisations, our direct application of belief propagation to carry out the computation has a practical runtime that is comparable to \gls{mcgo}. Furthermore, on some networks, \gls{JFC} is faster than \gls{mcgo}, probably due to having a lower overhead and being better able to leverage the optimized code in the \gls{pgmpy} library for the bulk of the computation.


\section{Case Study in Model Selection}
Although allowing for a simpler implementation that can leverage existing library implementations of the junction tree algorithm for most of the computation is a satisfactory result by itself, recall that the original motivation of \gls{JFC} is to compute a wider range of divergences between graphical models.
Therefore, in order to motivate the need of using divergences other than the \gls{kl} divergence, we now present a case study on the application of computing divergences between \glspl{bn} for the problem of model selection, a problem that the \gls{kl} divergence is normally well suited for.

Consider a scientist who, in an attempt to model a natural phenomenon that they have samples from, constructs 2 candidate \glspl{bn}, $A$ and $B$. They then wish to determine, using the samples, which candidate model is a better representation of the phenomenon they wish to model. One way to do this, is to estimate a new \gls{bn}, $E$, from the samples and compute the divergence between $E$ and the candidate models.

In order to recreate this scenario synthetically, we use the \gls{bn} \texttt{sachs} from the bnlearn repository~\cite{scutari2010} as the ``phenomenon'' the scientist wishes to model. The scientist's ``candidate models'' are then constructed by removing edges from \texttt{sachs} and marginalising the \glspl{cpt} according to~\cite{choi2005}. Further details regarding the construction of $A$ and $B$ can be found in Appendix~\ref{apx:construct}.

Sampling $100000$ samples from \texttt{sachs}, we then learn \gls{bn} $E$ from these samples using the constraint-based structure learner in \textit{pgmpy} and \textit{maximum likelihood estimation} with Laplace smoothing for learning the parameters of $E$.
The use of a smoothing technique is to ensure that the \gls{kl} divergence is defined.
We then compute the Hellinger and \gls{kl} divergence between the candidate models and the estimated model: $D(\text{A} \mid \mid  \text{E})$ and $D(\text{B} \mid \mid  \text{E})$. We repeat the experiment $5$ times, with different random samples from \texttt{sachs}.

\begin{table}
  \centering
  \begin{tabularx}{\columnwidth}{lXXXX}
    \toprule
    \multirow{2}{*}{run}
    & \multicolumn{2}{c}{Kullback-Leibler}
    & \multicolumn{2}{c}{Hellinger}\\
    \cmidrule{2-5}
    & $A \mid\mid E$ & $B \mid \mid E$ & $A,E$ & $B,E$\\
    \midrule
    1&\textbf{0.4021}&0.5169&0.3027&\textbf{0.2915}\\
    2&\textbf{0.3993}&0.5182&0.3009&\textbf{0.2895}\\
    3&\textbf{0.3979}&0.5234&0.3014&\textbf{0.2906}\\
    4&\textbf{0.4018}&0.5219&0.3022&\textbf{0.2904}\\
    5&\textbf{0.3996}&0.5275&0.3018&\textbf{0.2908}\\
    \bottomrule
  \end{tabularx}
  \caption{Divergence between the candidate models and a Bayesian network estimated from randomly sampled datasets of size $10000$. Lower numbers indicate a better fit and are bold.}\label{tbl:expr-model-sel}
\end{table}

From the results in Table~\ref{tbl:expr-model-sel}, we can observe that the \gls{kl} divergence indicates that $A$ is the \gls{bn} closest to $E$ and that the scientist should choose $A$, while the Hellinger distance indicates the opposite, choosing $B$ instead. With this discrepancy, the question then is, which candidate model, $A$ or $B$, is actually the closer approximation to the actual phenomenon, and therefore, which divergence is ``correct''.

Since, for the purpose of this case study, we already have the true model of the ``phenomenon'' we are modelling, we can just compute the divergence between our candidate models and \texttt{sachs} to get an answer.
\begin{table}
  \centering
  \begin{tabularx}{\columnwidth}{lXX}
    \toprule
    & \texttt{sachs}$\mid \mid A$ & \texttt{sachs}$ \mid \mid B$\\
    \midrule
    Kullback-Leibler\ \ \ & 0.3687 & \textbf{0.3090}\\
    Hellinger & 0.3013 & \textbf{0.2921}\\
    \bottomrule
  \end{tabularx}
  \caption{Divergence between the candidate models and the original Bayesian network \texttt{sachs}. Lower numbers indicate a better fit. Lowest divergences are bold.}\label{tbl:expr-dist-ori}
\end{table}
From Table~\ref{tbl:expr-dist-ori}, we can observe that when computing the divergence between \texttt{sachs} and the candidate models, both divergences agree that $B$ is closer to \texttt{sachs}. Consequently, in our case study, our scientist would have chosen the incorrect model if they only used the \gls{kl} divergence in their experiment.

Of course, it might be possible to avoid such a scenario if a different smoothing technique is used to learn the parameters of $E$. However, the use of multiple divergences is still needed in order for the scientist to even be aware of possible issues in the smoothing technique used in the first place. In general, the main takeaway from this example should be that, one must not be over-reliant on just a single divergence, and that the use of a wide array of divergences can be helpful in avoiding mistakes in model selection.


\section{Conclusion}\label{sec:conc}
In conclusion, we showed how computing the functional $\func$, and therefore the $\ab$-divergence, between 2 \glspl{dm} is equivalent to belief propagation on a junction tree/forest with a set of specific initial factors defined based on how the MLE of \glspl{dm} decomposes the functional $\func$.
The result is a method
with complexity exponential to the treewidth of the computation graph $\graphH$ of these models.
Therefore, the proposed method is more efficient than computing the $\func$ between the \glspl{dm} directly unless $\graphH$ is a fully saturated graph.

One advantage of \gls{JFC} is that it can be easily implemented in any environment that has a pre-existing implementation of the junction tree algorithm. Furthermore, since \gls{JFC} can compute the general functional $\func$ between 2 \glspl{dm}, it can compute, or approximate, other divergences or functionals, and not just the $\ab$-divergence.

However, recall that in order to obtain the computation graph $\graphH$, we take the graph union of the two \glspl{dm} we wish to compute the divergence between, and triangulate the resulting graph union to form a chordal computation graph.
Therefore, one potential area of concern is the possibility for the triangulation step to produce a computation graph $\graphH$ that has a large treewidth.
Due to the complexity being exponential to the treewidth of $\graphH$, this will result in the exact value of $\func$ taking a long time to compute.
Therefore, one avenue of further research is doing away with the requirement that the computation graph $\graphH$ has to be a chordal graph in exchange for an approximation of $\func$ instead of an exact computation.
In principle, this can be done by not triangulating $\graphP \cup \graphQ$, and instead running approximate inference algorithms on the graph $\graphP \cup \graphQ$ using the set of factors $\Psi$ defined in Theorem~\ref{thm:sp-compute}.

Furthermore, throughout this work, we only considered estimating the divergence between 2 discrete distributions.
Therefore, more work is needed to investigate how one might extend this approach for divergence estimation to numeric or even mixed type data.

Additionally, in this work we were only concerned with computing the divergence between the joint distributions of 2 \glspl{dm}.
However, in practice, it is common to encounter situations where one might want to compute the divergence between 2 conditional distributions.
Therefore, more work is needed to investigate this particular problem either by extending the current work or via some new method that only draw inspiration from the current work.



\section*{Acknowledgments}
This research has been funded by the Federal Ministry of Education and
Research of Germany and the state of North-Rhine Westphalia as part of
the Lamarr-Institute for Machine Learning and Artificial Intelligence,
LAMARR22B, the Australian Research Council under award DP210100072, as
well as the Australian Government Research Training Program (RTP)
Scholarship.


\bibliographystyle{aaai23}
\bibliography{ref,ref_other,new}

\clearpage
\appendix
\section{Proof of Theorem \ref{thm:ab-div-f-expr}}
\noindent
  When $\alpha,\beta,\alpha+\beta\neq0$:
  \begin{align*}
    &D_{AB}^{\alpha,\beta}(\pr,\qr)\\
    &=-\frac{1}{\alpha\beta}\left(
      \sum_{\xval\in\dom}\pr(\vec{x})^{\alpha}\qr(\vec{x})^{\beta} -
      \frac{\alpha}{\alpha + \beta} \sum_{\xval\in\dom}\pr(\vec{x})^{\alpha + \beta} -\right.\\
    &\left.\qquad\qquad\qquad
      \frac{\beta}{\alpha + \beta} \sum_{\xval\in\dom}\qr(\vec{x})^{\alpha + \beta}
      \right)\\
    &=-\frac{1}{\alpha\beta}\left(
      f_{2}(\pr,\qr;\alpha,\beta) -
      \frac{\alpha}{\alpha + \beta} f_{2}(\pr,\qr;\alpha+\beta,0) -\right.\\
    &\left.\qquad\qquad\qquad
      \frac{\beta}{\alpha + \beta} f_{2}(\pr,\qr;0,\alpha+\beta)
      \right)
  \end{align*}
  When $\alpha\neq0,\beta=0$:
  \begin{align*}
    &D_{AB}^{\alpha,\beta}(\pr||\qr)\\
    &=\frac{1}{\alpha^{2}}\Biggl(
      \sum_{\xval\in\dom}\pr(\vec{x})^{\alpha}
      \ln\frac{\pr(\vec{x})^{\alpha}}{\qr(\vec{x})^{\alpha}} -\\
      &\qquad\qquad \sum_{\xval\in\dom}\pr(\vec{x})^{\alpha} +
      \sum_{\xval\in\dom}\qr(\vec{x})^{\alpha}
      \Biggr)\\
    &=\frac{
      f_{3}(\pr,\qr;\alpha,0,\alpha,-\alpha) -
      f_{2}(\pr,\qr;\alpha,0) +
      f_{2}(\pr,\qr;0,\alpha)
      }{\alpha^{2}}
  \end{align*}
  When $\alpha=-\beta\neq0$:
  \begin{align*}
    &D_{AB}^{\alpha,\beta}(\pr||\qr)\\
    &=\frac{1}{\alpha^{2}} \left(
      \sum_{\xval\in\dom}\ln\frac{\qr(\vec{x})^{\alpha}}{\pr(\vec{x})^{\alpha}} +
      \sum_{\xval\in\dom}\left(
        \frac{\qr(\vec{x})^{\alpha}}{\pr(\vec{x})^{\alpha}}\right)^{-1} -
      \sum_{\xval\in\dom}1
      \right)\\
    &=\frac{f_{3}(\pr,\qr;0,0,-\alpha,\alpha) +
      f_{3}(\pr,\qr;0,0,\alpha,-\alpha) -
      |\dom|
      }{\alpha^{2}}
  \end{align*}
  When $\alpha=0,\beta\neq0$:
  \begin{align*}
    &D_{AB}^{\alpha,\beta}(\pr||\qr)\\
    &=\frac{1}{\beta^{2}}\Biggl(
      \sum_{\xval\in\dom}\qr(\vec{x})^{\beta}
      \ln\frac{\qr(\vec{x})^{\beta}}{\pr(\vec{x})^{\beta}} - \\
      &\qquad\qquad \sum_{\xval\in\dom}\qr(\vec{x})^{\beta} + \sum_{\xval\in\dom}\pr(\vec{x})^{\beta}
    \Biggr)\\
    &=\frac{
      f_{3}(\pr,\qr;0,\beta,-\beta,\beta) -
      f_{2}(\pr,\qr;0,\beta) +
      f_{2}(\pr,\qr;\beta,0)
      }{\beta^{2}}
  \end{align*}
  When $\alpha,\beta=0$:
  \begin{align*}
    D_{AB}^{\alpha,\beta}(\pr||\qr)
    &=\frac{1}{2}(\sum_{\xval\in\dom}\ln \pr(\vec{x}) -
      \sum_{\xval\in\dom}\ln \qr(\vec{x}))^{2}\\
    &=f_{1}(\pr,\qr)
  \end{align*}
  \qed


\section{Proof of Theorem \ref{thm:f1-timecmplx}}
\noindent
  Recall the $\alpha\beta$-divergence when $\alpha,\beta=0$:
  \begin{align*}
    &D_{AB}^{0,0}(\pr||\qr)\\
    &=\frac{1}{2}\sum_{\vec{x}\in\dom}
      {\left(\log \pr(\vec{x}) -
      \log \qr(\vec{x})\right)}^{2}\\
    &=\frac{1}{2}\sum_{\vec{x}\in\dom} \Bigl(
      {\log(\pr(\vec{x}))}^{2} +
      {\log(\qr(\vec{x}))}^{2}-\\
      &\qquad\qquad\quad 2 \left(\log \pr(\vec{x}) \log \qr(\vec{x})\right) \Bigr)
  \end{align*}
  We will first show how the sum over $\dom$ for the term $-2\log(\pr(\xval))\log(\qr(\xval))$ can be done while avoiding complexity exponential to $n$ (i.e.\ complexity linear to $|\dom|$). By the end of this process, we will then be able to observe that the complexity for computing the sum over $\dom$ for $\log(\pr(\xval))^{2}$ and $\log(\qr(\xval))^{2}$ are equivalent to the complexity for computing the sum over $\dom$ for $\log(\pr(\xval))\log(\qr(\xval))$ in big-O notation.

  Substituting the MLE for \glspl{dm} into $\log(\pr(\xval))\log(\qr(\xval))$, we get:
  \begin{align*}
    \sum_{\vec{x}\in\dom}
    &\log \pr(\vec{x}) \log \qr(\vec{x})\\
    =\sum_{\vec{x}\in\dom}
    &\left[
      \sum_{\cli \in \clis_{\pr}} \log \pr_{\cli}(\vec{x}) -
      \sum_{\sep \in \seps_{\pr}} \log \pr_{\sep}(\vec{x})
      \right]\\
    &\left[
      \sum_{\cli' \in \clis_{\qr}} \log \qr_{\cli'}(\vec{x}) -
      \sum_{\sep' \in \seps_{\qr}} \log \qr_{\sep'}(\vec{x})
      \right]\\
    =\sum_{\vec{x}\in\dom}
    &\left[\sum_{\cli\in\clis_{\pr}}\sum_{\cli'\in\clis_{\qr}}
      \log\pr_{\cli}(\vec{x}) \log\qr_{\cli'}(\vec{x}) -
    \right.\\
    &\quad \sum_{\cli\in\clis_{\pr}}\sum_{\sep'\in\sep_{\qr}}
      \log\pr_{\cli}(\vec{x}) \log\qr_{\sep'}(\vec{x}) -\\
    &\quad\sum_{\sep\in\seps_{\pr}}\sum_{\cli'\in\clis_{\qr}}
      \log\pr_{\sep}(\vec{x}) \log\qr_{\cli'}(\vec{x}) +\\
    &\left.\quad\sum_{\sep\in\seps_{\pr}}\sum_{\sep'\in\seps_{\qr}}
      \log\pr_{\sep}(\vec{x}) \log\qr_{\sep'}(\vec{x})\right]
  \end{align*}
  All four terms in this sum can be computed by pushing/rearranging the sum over $|\dom|$ to be the inner-most sum and conducting basic variable elimination between 2 terms in the sum:
  \begin{align*}
    \intertext{$\forall (\mathcal{B},\mathcal{D}) \in \{(\clis_\pr, \clis_\qr), (\clis_\pr, \seps_\qr), (\seps_\pr,\clis_\qr), (\seps_\pr,\seps_\qr)\}$}
    &\sum_{\vec{x}\in\dom}\sum_{B\in\mathcal{B}}\sum_{D\in\mathcal{D}} \log\pr_{B}(\vec{x}) \log\qr_{D}(\vec{x})\\
    &=\sum_{B\in\mathcal{B}}\sum_{D\in\mathcal{D}} \text{VE}(\dom,\log\pr_{B}, \log\qr_{D})\\
    \intertext{where for $D,B\subseteq \xvars$:}
    &\text{VE}(\dom, f_{D}, f_{B})\\
    &=\begin{cases}
      \left( \sum_{\xval \in \dom_{D}} f_{D}(\xval) \right) \left( \sum_{\xval \in \dom_{B}} f_{B}(\xval) \right) &
      D \cap B = \emptyset\\
      \sum_{\xval \in \dom_{D}} f_{D}(\xval)  f_{B}(\xval) &
      D = B\\
      \sum_{\xval \in \dom_{D}} f_{D}(\xval)  \sum_{\xval' \in \dom_{B - D}} f_{B}(\xval') &
      B \subset D\\
      \sum_{\xval \in \dom_{B}} f_{B}(\xval)  \sum_{\xval' \in \dom_{D - B}} f_{D}(\xval') &
      \text{otherwise}
    \end{cases}
  \end{align*}
  The complexity of computing $\text{VE}(\dom, f_{D},f_{B})$ involves determining which case $B$ and $D$ satisfies which takes time linear to $\mathcal{O}(\text{max}(|B|,|D|))$, and carrying out the variable elimination step which takes time $\mathcal{O}(2^{\text{max}(|B|,|D|)})$. We also know that $|B|$ and $|D|$ are bounded by the largest clique size in chordal graphs $\graphP$, $\omega(\graphP)+1$, and $\graphQ$, $\omega(\graphQ)+1$ respectively. Therefore, computing $\text{VE}(\dom, f_{D},f_{B})$ takes complexity $\mathcal{O}(\omega2^{\omega+1})$, where $\omega = \text{max}(\omega(\graphP), \omega(\graphQ))$.

  Furthermore, when computing $\sum_{\xval\in\dom}\log \pr(\xval)\log\qr(\xval)$, $\text{VE}(\dom, f_{D},f_{B})$ is computed for $(|\clis_{\pr}| + |\seps_{\pr}|)(|\clis_{\qr}| + |\seps_{\qr}|)$ distinct $B$s and $D$s, resulting in a complexity of:
  \begin{align*}
    &\sum_{\xval\in\dom}\log(\pr(\xval))\log(\qr(\xval))\\
    &\quad\in\mathcal{O}\big(
    (|\clis_{\pr}| + |\seps_{\pr}|)
    (|\clis_{\qr}| + |\seps_{\qr}|)\omega 2^{\omega+1}\big)
  \end{align*}
  Since the number of cliques and separators in chordal graph $\graph$ is bounded by the number of vertices in $\graph$, they are also bounded by the number of random variables associated with the graph $\xvars$. In other words:
  \begin{gather*}
    \bigO(\clis) \in \bigO(n)\\
    \bigO(\seps) \in \bigO(n)
  \end{gather*}
  therefore:
  \begin{align*}
    &\bigO\big(
    (|\clis_{\pr}| + |\seps_{\pr}|)
    (|\clis_{\qr}| + |\seps_{\qr}|)\omega 2^{\omega+1}\big)\\
    &\quad\in \bigO\bigl({(2n)}^{2} \omega 2^{\omega+1}\bigr)
    \in \bigO\bigl({n}^{2} \omega 2^{\omega+1}\bigr)
  \end{align*}
  Using the same argumentation, the complexity of computing $\sum_{\xval\in\dom}\log(\pr(\xval))^{2}$ and $\sum_{\xval\in\dom}\log(\qr(\xval))^{2}$ is:
  \begin{align*}
    &\sum_{\xval\in\dom}\log(\pr(\xval))^{2}\\
    &\quad\in \mathcal{O}((|\clis_{\pr}| + |\seps_{\pr}|)^{2} \omega(\graphP) 2^{\omega(\graphP)+1})
    \in\bigO\bigr(n^{2}\omega 2^{\omega+1}\bigl)\\
    &\sum_{\xval\in\dom}\log(\qr(\xval))^{2}\\
    &\quad\in \mathcal{O}((|\clis_{\qr}| + |\seps_{\qr}|)^{2} \omega(\graphP) 2^{\omega(\graphP)+1})
    \in\bigO\bigr(n^{2}\omega 2^{\omega+1}\bigl)
  \end{align*}
  Therefore, the complexity of computing $D_{AB}^{0,0}(\pr||\qr)$ is:
  \begin{align*}
    D_{AB}^{0,0}(\pr||\qr)
    \in\bigO\bigr(n^{2}\omega 2^{\omega+1}\bigl)
  \end{align*}
  which implies that the functional $f_{1}$ can be computed naively while avoiding complexity exponential to the number of variables $n$.
  \qed


\section{Proof of Theorem \ref{thm:express-f2}}
Recall $\func$ from Definition~\ref{def:func} and $f_{2}$ from Theorem~\ref{thm:ab-div-f-expr}:
\begin{align*}
  &\func(\pr, \qr; g, h, g^{*}, h^{*}, L)\\
  &=\sum_{\xval\in\dom}
    [g[\pr](\xval)]\cdot [h[\qr](\xval)] \cdot
    L\left([g^{*}[\pr](\xval)] \cdot
    [h^{*}[\qr](\xval)]\right)\\
  &f_{2}(\pr, \qr; \alpha, \beta)
  =\sum_{\xval\in\dom}\pr(\xval)^{\alpha}\qr(\xval)^{\beta}
\end{align*}
First, set the parameters of $\func$ to be:
\begin{gather*}
  \begin{aligned}
    &h[\pr](\xval) = { \pr }(\xval)^{\alpha}
    &g[\qr](\xval) = { \qr }(\xval)^{\beta}\\
  \end{aligned}\\
  h^{*}[ \pr_{{B}} ] =
  b^{1/2(|\clis(\graphP)| - |\clis(\graphP({B}))| + 1)}\\
  g^{*}[ \qr_{{D}} ] =
  b^{1/2(|\clis(\graphQ)| - |\clis(\graphQ({D}))| + 1)}\\
  L(x) = \log_{b}(x)
\end{gather*}
where $\graph(S)$ for $S \subset \xvars$ is the induced subgraph of $\graph$ over the variables in $S$; $B,D\subset\xvars$ such that $\graphP(B)$ and $\graphQ(D)$ are chordal graphs where $\clis(\graphP) = \clis(\graphP(B)) \cup \clis(\graphP(X - B))$ and $\clis(\graphQ) = \clis(\graphQ(D)) \cup \clis(\graphQ(X - D))$; and the logarithmic base $b$ is any real number.

This specific parameterisation of $f^{*}$ and $g^{*}$ ensures that they satisfy the requirement in Equation~\ref{eq:func-multi-prop} of Definition~\ref{def:func} while also removing the log term from the functional $\func$. Using these parameters we get:
\begin{align*}
  &\func(\pr, \qr)\\
  &= \sum_{\xval\in\dom}
  { \pr(\xval) }^{\alpha} {\qr(\xval)}^{\beta}\log_{b}\left(
  b^{1/2(|\clis(\graphP)| - |\clis(\graphP({\xvars}))| + 1)}\right.\\
  &\qquad\qquad\qquad\qquad\qquad\qquad\left.
    b^{1/2(|\clis(\graphQ)| - |\clis(\graphQ({\xvars}))| + 1)}
    \right)\\
  &= \sum_{\xval\in\dom}
    { \pr(\xval) }^{\alpha} {\qr(\xval)}^{\beta}\log_{b}\left(b^{1}\right)\\
   &= \sum_{\xval\in\dom}
    { \pr(\xval) }^{\alpha} {\qr(\xval)}^{\beta}
\end{align*}
\qed

\section{Proof of Theorem \ref{thm:express-f3}}
Recall $\func$ from Definition~\ref{def:func} and $f_{3}$ from Theorem~\ref{thm:ab-div-f-expr}:
\begin{align*}
  &\func(\pr, \qr; g, h, g^{*}, h^{*}, L)\\
  &=\sum_{\xval\in\dom}
    [g[\pr](\xval)]\cdot [h[\qr](\xval)] \cdot
    L\left([g^{*}[\pr](\xval)] \cdot
    [h^{*}[\qr](\xval)]\right)\\
  &f_{3}(\pr, \qr; \alpha, \beta,c,d)=\sum_{\xval\in\dom}\pr(\xval)^{\alpha}\qr(\xval)^{\beta}
    \log(\pr(\xval)^{c}\qr(\xval)^{d})
\end{align*}
Therefore $\func=f_{3}$ given the following parameters for $\func$:
\begin{gather*}
  \begin{aligned}
    &h[\pr](x) = \pr(x)^{a}
    &h^{*}[\pr](x) = \pr(x)^{c}\\
    &g[\qr](x) = \qr(x)^{b}
    &g^{*}[\qr](x) = \qr(x)^{d}\\
  \end{aligned}\\
  L(x) = \log x
\end{gather*}
\qed

\section{Proof of Remark~\ref{prop:naive-complex}}
  The complexity of computing Equation~\ref{eq:func-est-plug-in} directly is exponential with respect to the product of the cardinality of each sum in the 2 nested sums of
  Equation~\ref{eq:func-est-plug-in}$\left( |\clis(\graph_{\pr})| + |\clis(\graph_{\qr})|\right)\text{Dom}(\xvars)$.
  Since the lower bound of the number of cliques in $\graph_{\pr}$ and
  $\graph_{\qr}$ is $1$, the lower bound complexity of computing Equation~\ref{eq:func-est-plug-in} is in
  $\Omega(2^{n})$.
  Therefore, the lower bound complexity of naively computing
  Equation~\ref{eq:func-est-plug-in} is exponential with respect to
  the number of variables in \glspl{dm} $\modelP$ and
  $\modelQ$.
  \qed


\section{Proof of Theorem \ref{thm:sp-eq}}
  We first show that the equivalence in Equation~\ref{eq:thm-sp-eq-d}
  holds for $\pr$.
  This will then imply that the equivalence holds for $\qr$ as well since the exact same
  steps can be used for showing that the equivalence holds for $\qr$.

  Starting from the left hand side of Equation~\ref{eq:thm-sp-eq-d},
  we can split the innermost sum into 2 sums, one over the set
  $\dom_{\alpha(C)-C}$ and $\dom_{X-\alpha(C)}$, then move the sum over
  $\dom_{\alpha(C)-C}$ outside and merge it with the sum over $\dom_{C}$:
  \begin{align*}
    &\sum_{C\in\clis_{\pr}} \sum_{\xval_{C}\in\dom_{C}}
    L\left(g^{*}\left[\prjt_{C}\right](\xval_{C})\right)
    \SP_{C}(\xval_{C})\\
    &=\sum_{C\in\clis_{\pr}} \sum_{\xval_{C}\in\dom_{C}}
      L\left(g^{*}\left[\prjt_{C}\right](\xval_{C})\right)\\
    &\qquad \sum_{\xval\in\dom_{X-C}}
      \big(g\left[\pr\right](\xval_{C},\xval)\big)
      \big(h\left[\qr\right](\xval_{C},\xval)\big)\\
    &=\sum_{C\in\clis_{\pr}} \sum_{\xval_{C}\in\dom_{C}}
      L\left(g^{*}\left[\prjt_{C}\right](\xval_{C})\right)\\
      &\qquad\sum_{\xval\in\dom_{X-\alpha(C)+\alpha(C)-C}}
      \big(g\left[\pr\right](\xval_{C},\xval)\big)
      \big(h\left[\qr\right](\xval_{C},\xval)\big)\\
    &=\sum_{C\in\clis_{\pr}} \sum_{\xval_{C}\in\dom_{C}}
      \sum_{\xval'\in\dom_{\alpha(C)-C}}
      L\left(g^{*}\left[\prjt_{C}\right]
      (\xval_{C},\xval')\right)\\
      &\qquad \sum_{\xval\in\dom_{X-\alpha(C)}}
      \big(g\left[\pr\right](\xval_{C},\xval,\xval')\big)
      \big(h\left[\qr\right](\xval_{C},\xval,\xval')\big)\\
    &=\sum_{C\in\clis_{\pr}} \sum_{\xval_{\alpha(C)}\in\dom_{\alpha(C)}}
      L\left(g^{*}\left[\prjt_{C}\right](\xval_{\alpha(C)})\right)\\
      &\qquad\sum_{\xval\in\dom_{X-\alpha(C)}}
      \big(g\left[\pr\right](\xval_{\alpha(C)},\xval)\big)
      \big(h\left[\qr\right](\xval_{\alpha(C)},\xval)\big)\\
    &=\sum_{C\in\clis_{\pr}} \sum_{\xval_{\alpha(C)}\in\dom_{\alpha(C)}}
      L\left(g^{*}\left[\prjt_{C}\right](\xval_{\alpha(C)})\right)
      \SP_{\alpha(C)}(\xval_{\alpha(C)})
  \end{align*}
  \qed


\section{Proof of Theorem \ref{thm:sp-compute}}
  By Definition~\ref{def:cli-map}, there exist a mapping, $\alpha$,
  from the maximal cliques in $\modelP$ and $\modelQ$ to a maximal
  clique in the computation graph $\graphH$.
  Therefore, each element in the factor set $\Phi$ can be mapped to a
  maximal clique in $\graphH$ as well.

  The existence of this mapping allows us to run a junction tree
  algorithm over the junction tree of graph $\graphH$,
  $\mathcal{T}=(\clis,\seps)$, using $\Phi$ as factors.

  Recall from Section~\ref{sec:back-jta} that given a chordal graph
  $\graphH$ and a set of factors $\Phi=\{\phi_{i}\}$, the Junction Tree
  Algorithm provides the following belief for each maximal clique in the
  Junction Tree, $\mathcal{T}=(\clis,\seps)$, of the chordal graph
  $\graphH$:
  \begin{align*}
    \forall C \in \clis : \beta_{C}(\xval_{C}) = \sum_{\xval \in \dom_{\xvar-C}}
    \prod_{\phi \in \Phi} \phi(\xval, \xval_{C})
  \end{align*}
  By using the set of factors $\Phi$ defined in
  Theorem~\ref{thm:sp-compute}, we directly obtain the following by
  simple substitution:
  \begin{align*}
    \forall C \in \clis :\\
    \beta_{C}(\xval_{C})
    &= \sum_{\xval \in \dom_{\xvar-C}}
    \prod_{C_{\pr} \in \clis\left(\graph_{\pr}\right)}
    g\left[\prejt_{C_{\pr}}\right](\xval,\xval_{C})\\
    &\qquad\qquad\quad\prod_{C_\qr \in \clis\left(\graph_{\qr}\right)}
      h\left[\qrejt_{C_{\qr}}\right](\xval,\xval_{C})\\
    &=\SP_{C}(\xval_{C})
  \end{align*}
  \qed



\section{Proof of Theorem \ref{thm:jfc-compute}}
\noindent
  Recall from Equation~\ref{eq:sp},
  $\forall C \in \clis(\graph_\pr,\graph_\qr)$:
  \begin{align*}
    \SP_{\alpha(C)}(\xval_{\alpha(C)})
   &=\sum_{\xval \in \dom_{\xvar-\alpha(C)}}
     \left[
     \prod_{C_\pr \in \clis_{\pr}}
     g\left[\prejt_{C_\pr}\right](\xval_{\alpha(C)},\xval)
     \right]\\
     &\qquad\qquad\qquad\left[
     \prod_{C_\qr \in \clis_{\qr}}
     h\left[\qrejt_{C_\qr}\right](\xval_{\alpha(C)}\xval)
     \right]
  \end{align*}
  Following the usual steps to initiate the junction tree algorithm,
  we define the set of factors, $\Phi$, to use in the junction tree
  algorithm as such:
  \begin{equation*}
    \Phi =
    \left\{g\left[\prejt_{C}\right] : C \in \clis_{\pr}\right\}
    \bigcup
    \left\{h\left[\qrejt_{C}\right] : C \in \clis_{\qr}\right\}
  \end{equation*}
  As usual at the start of the junction tree algorithm, each factor in
  $\Phi$ is assigned to a maximal clique in the junction tree, or in this
  case junction forest, of the computation graph $\graphH$.
  These assignments will use the clique mapping function from
  Definition~\ref{def:cli-map}, $\alpha$, and the factors assigned into
  each maximal clique in $\graph$ are multiplied together into the
  initial belief for each clique $\psi_{i}$:
  \begin{align*}
    \forall C \in &\clis(\graphH):\\
    &\psi_{C} =
    \prod_{C_\pr \in A_{C}(\clis(\graph_\pr))}
    g\left[\prejt_{C_\pr}\right]
    \prod_{C_\qr \in A_{C}(\clis(\graph_\qr))}
    h\left[\prejt_{C_\qr}\right]
  \end{align*}
  Then $\SP$ can be re-expressed in terms of these initial clique
  beliefs $\psi$:
  \begin{align*}
    SP_{\alpha(C)}(\xval_{\alpha(C)})
    &=\sum_{\xval\in\dom_{\xvar-\alpha(C)}}
      \prod_{C_\graphH \in \clis(\graphH)}
      \psi_{C_\graphH}(\xval_{\alpha(C)},\xval)
  \end{align*}
  Due to the independence between maximal cliques of different
  junction trees in the junction forest $\jt$, we can split the sum in
  $SP$ into a product of sum-products over each junction tree in
  $\jt$:
  \begin{align*}
    &SP_{\alpha(C)}(\xval_{\alpha(C)})\\
    &=\sum_{\xval\in\dom_{\xvar-\alpha(C)}}
      \prod_{C_\graphH \in \clis(\graphH)}
      \psi_{C_\graphH}(\xval_{\alpha(C)},\xval)\\
    &=\sum_{\xval\in\dom_{\xvar-\alpha(c)}}
      \prod_{T\in\jt} \prod_{C_\graphH\in\clis(T)}
      \psi_{C_\graphH}(\xval_{\alpha(C)},\xval)\\
    &=\sum_{\xval_{T_{1}} \in \dom_{\clis(\jt_{1})}} \cdots
      \sum_{\xval_{\tau(C)} \in \dom_{\clis(\tau(C))-\alpha(C)}} \cdots
      \sum_{\xval_{T_{|\jt|}} \in \dom_{\clis\left(T_{|\jt|}\right)}}\\
      &\qquad\prod_{T\in\jt} \prod_{C_\graphH\in\clis(T)}
      \psi_{C_\graphH}(\xval_{T_{1}}, \ldots,\xval_{\tau(c)},\ldots,\xval_{T_{|\jt|}})\\
    &=\left(\sum_{\xval_{\tau(C)}\in\dom_{\clis(\tau(C))-\alpha(C)}}
      \prod_{C_\graphH\in\clis(\tau(C))}\psi_{C_\graphH}(\xval_{\tau(C)})\right)\\
      &\qquad\prod_{T\in\jt-\tau(C)} \left(
      \sum_{\xval_{T}\in\dom_{\clis(T)}}
      \prod_{C_\graphH\in\clis(T)}\psi_{C_\graphH}(\xval_{T})
      \right)\\
    &=\beta_{\alpha(C)}(\xval_{\alpha(C)})\prod_{T\in\jt-\tau(C)}
      \sum_{\xval\in\dom_{C(T)}} \beta_{C(T)}(\xval_{\alpha(C)},\xval)
  \end{align*}
  \qed


\section{Constructing candidate models for case study in model selection}\label{apx:construct}
The candidate model $A$ is obtained by removing the following directed edges from \texttt{sachs}:
\begin{itemize}
  \item \texttt{PKA} $\to$ \texttt{Raf}
  \item \texttt{PKC} $\to$ \texttt{PKA}
  \item \texttt{Plcg} $\to$ \texttt{PIP3}
\end{itemize}
and the candidate model $B$ is obtained by removing the following directed edges from \texttt{sachs}:
\begin{itemize}
  \item \texttt{PKC} $\to$ \texttt{Raf}
  \item \texttt{PKC} $\to$ \texttt{Mek}
  \item \texttt{PKA} $\to$ \texttt{Mek}
\end{itemize}
On the deletion of an edge $Y\to X$, the \gls{cpt} of $X$ is marginalised as follows~\cite{choi2005} where $\bm{Z}=\text{parents}(X) \setminus Y$:
\begin{align*}
  \pr_{X | \bm{Z}}'(x|\bm{z})
  \vcentcolon= \sum_{y\in\text{Dom}(Y)} \pr_{X | Y,\bm{Z}}(x | y,\bm{z}) \pr_{Y}(y)
\end{align*}


\end{document}